\documentclass[journal]{IEEEtran}
\usepackage{amssymb} 
\usepackage{autobreak} 
\usepackage{booktabs}
\usepackage{cite}
\usepackage{marvosym}
\usepackage{times}
\usepackage{epsfig}
\usepackage{graphicx}
\usepackage{amssymb}
\usepackage{booktabs}
\usepackage{xcolor}
\usepackage{multirow}
\usepackage{graphics}

\ifCLASSINFOpdf
\else
   \usepackage[dvips]{graphicx}
\fi
\usepackage{url}

\usepackage{graphicx}

\begin{document}

\title{Perceptual Constancy Constrained Single Opinion Score Calibration for Image Quality Assessment}
\author{Lei Wang, Desen Yuan 
\thanks{The authors are with the School of Information and Communication Engineering, University of Electronic Science and Technology of China, Chengdu, 611731, China.}

}

\markboth{Journal of \LaTeX\ Class Files, Vol. 14, No. 8, August 2015}
{Shell \MakeLowercase{\textit{et al.}}: Bare Demo of IEEEtran.cls for IEEE Journals}
\maketitle

\begin{abstract}
In this paper, we propose a highly efficient method to estimate an image's mean opinion score (MOS) from a single opinion score (SOS). Assuming that each SOS is the observed sample of a normal distribution and the MOS is its unknown expectation, the MOS inference is formulated as a maximum likelihood estimation problem, where the perceptual correlation of pairwise images is considered in modeling the likelihood of SOS. More specifically, by means of the quality-aware representations learned from the self-supervised backbone, we introduce a learnable relative quality measure to predict the MOS difference between two images. Then, the current image's maximum likelihood estimation towards MOS is represented by the sum of another reference image's estimated MOS and their relative quality. Ideally, no matter which image is selected as the reference, the MOS of the current image should remain unchanged, which is termed perceptual cons
tancy constrained calibration (PC3). Finally, we alternatively optimize the relative quality measure's parameter and the current image's estimated MOS via backpropagation and Newton's method respectively. Experiments show that the proposed method is efficient in calibrating the biased SOS and significantly improves IQA model learning when only SOSs are available. 
\end{abstract}


\IEEEpeerreviewmaketitle

\section{Introduction}
\label{sec:intro}

Image quality assessment (IQA) \cite{zhai2020perceptual,ma2017end,ma2021blind,wang2020blind,shao2013perceptual,zhang2021fine,xu2016multi,wu2020subjective} aims to enable computers to automatically predict the subjective quality of an image, which is consistent with the human perception. As the ground truth, the subjective quality label is the key to IQA model learning and evaluating. However, due to subjective preference differences, collecting reliable subjective quality labels is greatly challenging, especially for a large-scale image dataset with diverse visual content. 

The mean opinion score (MOS) \cite{streijl2016mean} is the most widely used subjective image quality label, which invites multiple human subjects and uses the mean value of their annotations as the image quality label. Given a large-scale image dataset, the MOS collection is usually achieved via crowdsourcing, whose annotation process is uncontrollable and easily results in noisy ratings. Then, a series of standardized outlier rejection and bias removal methods are developed by the international telecommunication union (ITU) recommendations including the P.910~\cite{installationsitu}, P.913~\cite{union2016methods}, BT.500~\cite{bt2020methodologies}. In~\cite{li2020simple}, Li \textit{et al.} further generalizes P.913's raw opinion scores screening procedure by joint modeling the subject bias and inconsistency. Despite successfully suppressing subjective preference differences, existing screening methods require multiple subjects in measuring the consistency of their ratings, which is very expensive and time-consuming.

To address this issue, in this paper, we propose a perceptual constancy constrained calibration (PC3) method, which could estimate the MOS from the single opinion score (SOS). Unlike the bias and inconsistency analysis across multiple subjects \cite{installationsitu,union2016methods,bt2020methodologies,li2020simple}, 
we model the SOS of an image with the normal distribution, whose unknown expectation is our desired MOS. In this way, the MOS estimation is formulated as a maximum likelihood estimation problem, which tries to find the distribution parameters to fit the observed SOS best. More specifically, we consider the perceptual correlation of pairwise images in modeling the likelihood of the SOS and represent the estimated MOS of an image as the sum of another reference image's estimated MOS and a learnable relative quality measure, which leverages the quality-aware representations learned from the self-supervised backbone \cite{madhusudana2022image} and predicts the MOS difference between two images. Ideally, no matter which image is selected as the reference, the MOS of the current image should remain unchanged. Finally, we alternatively optimize the relative quality measure's parameter and the current image's estimated MOS via backpropagation and Newton's method respectively.

This simple and effective method is suitable for subjective quality label post-processing when only single opinion scores are available, which significantly reduces the cost of MOS collection. Experiments on four popular IQA datasets show that the proposed method is efficient in calibrating the biased SOS and significantly improves the performance of the IQA model when only the SOSs are available for training.

\begin{figure*}
	\centering
	\includegraphics[width=\linewidth]{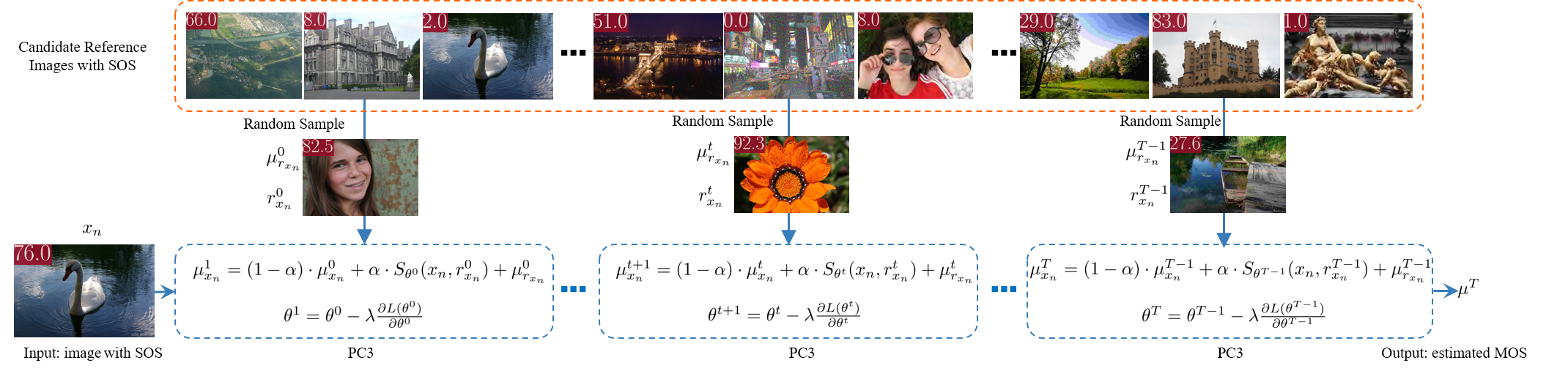} 	
 \caption{The overall framework of the proposed method. The input includes an image $x_n$ and its SOS $y_n$. We calibrate $y_n$ by iteratively updating the estimated MOS $\mu_{x_n}$ via our PC3, which requires a randomly selected reference image.}
	\label{fig:arch}
\end{figure*}

\section{Related Work} 
\noindent{\bf Image quality label collection.}  
Human decision-making presents uncertainty \cite{uncertainty}, which brings a great challenge for image quality label collection. To reduce external interference as much as possible, ITU recommends a highly controlled environment for the subjective tests. In BT.500~\cite{bt2020methodologies}, the number of subjects, viewing conditions, test materials, presentation order, and time are all standardized to collect the subjects' ratings. The brightness and contrast of the display devices and the standard
definition of test materials are also discussed in \cite{site2subjective}.

Due to limited resources, this highly controlled environment is hard to apply to multiple subjects' parallel ratings. In practice, more and more subjective tests turn to the crowdsourcing platform \cite{fang2020perceptual}, especially for large-scale image datasets, which invites multiple internet workers to conduct uncontrolled subjective experiments. Unsurprisingly, more noise would be collected in these uncontrolled subjective experiments.

\noindent{\bf Post-processing for unreliable ratings.} 
To clean the raw opinion scores, there are also several post-processing methods developed to detect and eliminate unreliable ratings \cite{zaric2011vcl}. To suppress the subjective preference difference of individuals, it is common practice to average the raw opinion scores from multiple subjects, yielding a MOS per stimulus. Standardized recommendations include more advanced corrective mechanisms to compensate for the influence of test subjects, as well as criteria for determining MOS confidence intervals. 

ITU-R BT.500~\cite{bt2020methodologies} Recommendation defines the procedure that counts the number of occurrences when a subject's opinion score deviates by a few sigmas (i.e. standard deviation) and rejects the subject if the occurrences are more than a fraction. The recommendation also specifies the method for calculating the confidence interval (ITU-R BT.500 Section A1-2.2.1). ITU and researchers have proposed a number of crowd-sourced data processing standards~\cite{bt2020methodologies,installationsitu,union2016methods} to eliminate subjective bias in MOS. These methods mainly rely on inconsistency analysis across multiple subjects, which is expensive, time-consuming, and difficult to apply to a single opinion score.

\section{Proposed Method}\label{sec:pl}
As shown in Fig. \ref{fig:arch}, we propose an iterative update framework for MOS estimation. The inputs include the image and its SOS, and the output is the estimated MOS. In each update step, we randomly sample a reference image with an estimated MOS and update the input image's MOS by measuring their relative quality. Let $y$ denote the SOS of an image $x$, and $p(y|x;\mu)$ denote the likelihood of SOS with the unknown expectation $\mu_x$, which is our desired MOS. Since there are no additional subject ratings for $x$, we try to further utilize the perceptual correlation of multiple images for SOS calibration, which reformulates the likelihood of SOS as
\begin{equation} 
\begin{aligned}
    p(y|x;\mu_{x})=&\mathrm{E}_{r_x|x}[p(y|x,{r_x},\mu_{r_x};\mu_{x})]\\
    =&\sum_{m=1}^M p[y|x,{r},\mu_{r_x(m)};\mu_{x}]p[r_x(m)|x]
\end{aligned}
\label{eq-reform}
\end{equation}
where $r_x(m)$ is the $m$th reference image of $x$, $\mu_{r_x(m)}$ is its MOS, and $M$ is the total number of available reference images.

According to our perceptual constancy constraint, $p[y|x,{r_x},\mu_{r_x(m)};\mu_{x}]$ and $p[r_x(m)|x]$ are the same for any $m$, which results to
\begin{equation} 
    p(y|x;\mu_{x})\propto p(y|x,{r_x},\mu_{r_x};\mu_{x}),
\end{equation}
where the likelihood of SOS is transformed from the conditional probability of a single image to the joint conditional probability of pairwise images.

In this context, we assume that $p(y|x,{r_x},\mu_{r_x};\mu_{x})$ still follows the normal distribution and introduces a relative quality measure $S_{\theta}(x,r_x)$ to describe it, i.e., 
\begin{equation}
\label{eq:pyxs}
p(y|x,{r_x},\mu_{r_x};\mu_{x}) \sim  \mathcal{N}\left(\mu_{r_x}+S_{\theta}(x,r_x), {\sigma}^{2} \right),
\end{equation}
where $\sigma$ is the standard deviation of the distribution. $S_{\theta}(x,r_x)$ is a Siamese network, whose two branches share the same backbone, i.e., CONTRIQUE~\cite{madhusudana2022image}. The two branches extract the quality-aware features from $x$ and $r_x$. The difference between them is then fed to three fully connected layers to obtain their relative quality, where $\theta$ is the parameter of this Siamese network. 
  
Following previous settings, given a collection of single opinion scores $\{y_n\}^{N}_{n=1}$ from a subjective experiment, our goal is to solve the unknown parameters $(\theta, U)$ to fit the observed SOSs best, where $U=\{\mu_{x_n}\}^{N}_{n=1}$. 
This task could be formulated as a maximum likelihood estimation problem. The log-likelihood of all images could be represented by
\begin{equation}
\begin{aligned}  
   \ell(\theta,U)&=\log \prod_{n=1}^N p[y_{n} \mid  \mu_{r_{x_n}}+S_{\theta}(x_n,r_{x_n})] \\
   &=\sum_{n=1}^{N}\log p[y_{n} \mid  \mu_{r_{x_n}}+S_{\theta}(x_n,r_{x_n})]
\end{aligned}.
\end{equation} 

It is noted that the maximum-likelihood estimate is equivalent to the least-squares estimate for normal distribution \cite{charnes1976equivalence}. Then, we replace the log-likelihood based objective function with the following mean square error loss, i.e., 
\begin{equation} 
   \ell(\theta,U)  \cong \frac{1}{N} \sum_{n=1}^{N}[y_{n}-S_{\theta}(x_n,r_{x_n})-\mu_{r_{x_n}}]^{2} 
   \label{eq:mse}
\end{equation} 
where $\cong$ denotes equal with the omission of constant terms. Besides the supervision from $y_n$, our perceptual constancy constraint also requires that the MOS of the current image should remain unchanged with different reference images. So, the MOS estimation is converted into a least-squares estimate problem with equality constraint, i.e.,
\begin{equation}
\underset{\theta,U}{\arg \min }~\ell(\theta,U) , \text { s.t. }  \mu_{x_n}=S_{\theta}(x_n,r_{x_n})+\mu_{r_{x_n}}.
\end{equation}

To balance these two objectives, we unify them together with the Lagrangian function
\begin{equation} 
    L(\theta,U)=\ell(\theta,U) +\beta\cdot R(\theta,U),
\end{equation} 
where $R(\theta,U)=\frac{1}{N} \sum_{n=1}^{N}[\mu_{x_n}-S_{\theta}(x_n,r_{x_n})-\mu_{r_{x_n}}]^{2}$, $\beta$ is the Lagrangian multiplier. Then, the current image's MOS $\mu_{x_n}$ and the relative quality measure's parameter $\theta$ are solved via an alternative optimization strategy. MOS estimation is updated by the Newton-Raphson rule, i.e.,
\begin{equation}
\begin{aligned}
\mu_{x_n}^{t+1} \leftarrow& (1-\alpha) \cdot \mu_{x_n}^{t}+\alpha \cdot \Delta^t 
\\\Delta^t=& \mu_{x_n}^t-\frac{\partial L(\mu_{x_n}) / \partial \mu_{x_n}}{2\lambda  \partial^2 L(\mu_{x_n}) / \partial \mu_{x_n}^2}
\\=& S_{\theta^t}(x_n,r_{x_n}^t)+\mu_{r_{x_n}}^t
\end{aligned},
\end{equation}
where $t$ is the iteration step, and $\mu_{x_n}^0=y_n$. As shown in Fig. \ref{fig:arch}, $r_{x_n}^t$ is randomly sampled from all candidate reference images, which may change in every iteration. $\alpha$ is a manually set refresh rate. Meanwhile, $\theta$ is optimized via the back-propagation of the Siamese network, i.e.,
\begin{equation} 
\theta^{t+1}=\theta^{t}- \lambda\cdot\frac{\partial L(\theta)}{\partial \theta},  
\end{equation}
where $\lambda$ is the learning rate. It is noted that the initial relative quality measure is unreliable. In our alternative optimization process, we set a warm-up time of $T_{h}$, where the MOS estimation update only activates when $S_{\theta}(\cdot,\cdot)$ has completed $T_{h}$ times optimization, i.e.,
\begin{equation}
u^{t+1}_{x_n} =
\left\{
    \begin{array}{cc}
         (1-\alpha) \cdot \mu_{x_n}^{t}+\alpha \cdot \Delta^t, & t\geq T_{h} \\ 
        y_{n}, & \text{otherwise}
    \end{array}.
\right.
\end{equation}

\section{Experiments}\label{sec:exp}

\label{sec:experiments}

\subsection{Datasets and Settings}
\label{sec:nmt}

In our experiment, we investigate the performance of the proposed method on four popular IQA databases including the VCL \cite{zaric2011vcl}, TID2013 \cite{ponomarenko2015image}, KONIQ\cite{hosu2020koniq}, LIVEC\cite{ghadiyaram2015massive}, which cover both the synthetic and authentic distortions. Since the subjective annotations are different across different databases, we develop specific SOS settings in the following three cases. \textbf{1) VCL:} All raw opinion scores are available. The SOS is obtained by randomly sampling a raw opinion score across all subjects for each image; \textbf{2) TID \& LIVEC:} MOS and the corresponding standard deviation are available. The SOS is obtained by random sampling from a normal distribution, which uses the pairwise MOS and standard deviation as the distribution parameters; \textbf{3) KONIQ:} The empirical distribution of all ratings is available. The SOS is obtained by random sampling from the empirical distribution of each image's ratings. To avoid bias toward specific SOS choices, we repeat the random sampling 10 times across all databases and report the median results for comparison.

For all databases, the original MOS is used as the ground truth. We evaluate the accuracy of the estimated MOS with three widely used metrics including Pearson's Linear Correlation Coefficient (PLCC), Spearman's Rank Order Correlation Coefficient (SRCC), and the Mean Square Error (MSE). To the best of our knowledge, SOS calibration is rarely discussed in IQA research. When only a single opinion score is available, the existing post-processing or screening methods \cite{installationsitu,union2016methods,bt2020methodologies,li2020simple} are inapplicable. For comparison, we chose the SOS as the baseline and consider the performance of MOS collected from abundant subjects as the theoretical upper bound. 

\subsection{Implementation Details}
In our experiment, we investigate the performance of the proposed SOS calibration method and its effect on IQA model learning, which employs a representative deep IQA model NIMA~\cite{talebi2018nima}. For both the $S_\theta(x_n,r_{x_n})$ and NIMA, the input images are resized and center-cropped to 320, and the initial learning rate $\lambda$ is set to 0.0001 with the Adam optimizer. For our PC3, the additional hyperparameter settings include $\beta = 1/9$, $\alpha = 0.1$, $T_{h} = 1$. For NIMA, we randomly split (except for KONIQ which has an official split) each IQA database into non-overlapped training and testing sets. The random split is repeated 10 times and the median results are reported in the following.

\begin{figure}
	\centering
	\includegraphics[width=0.8\linewidth]{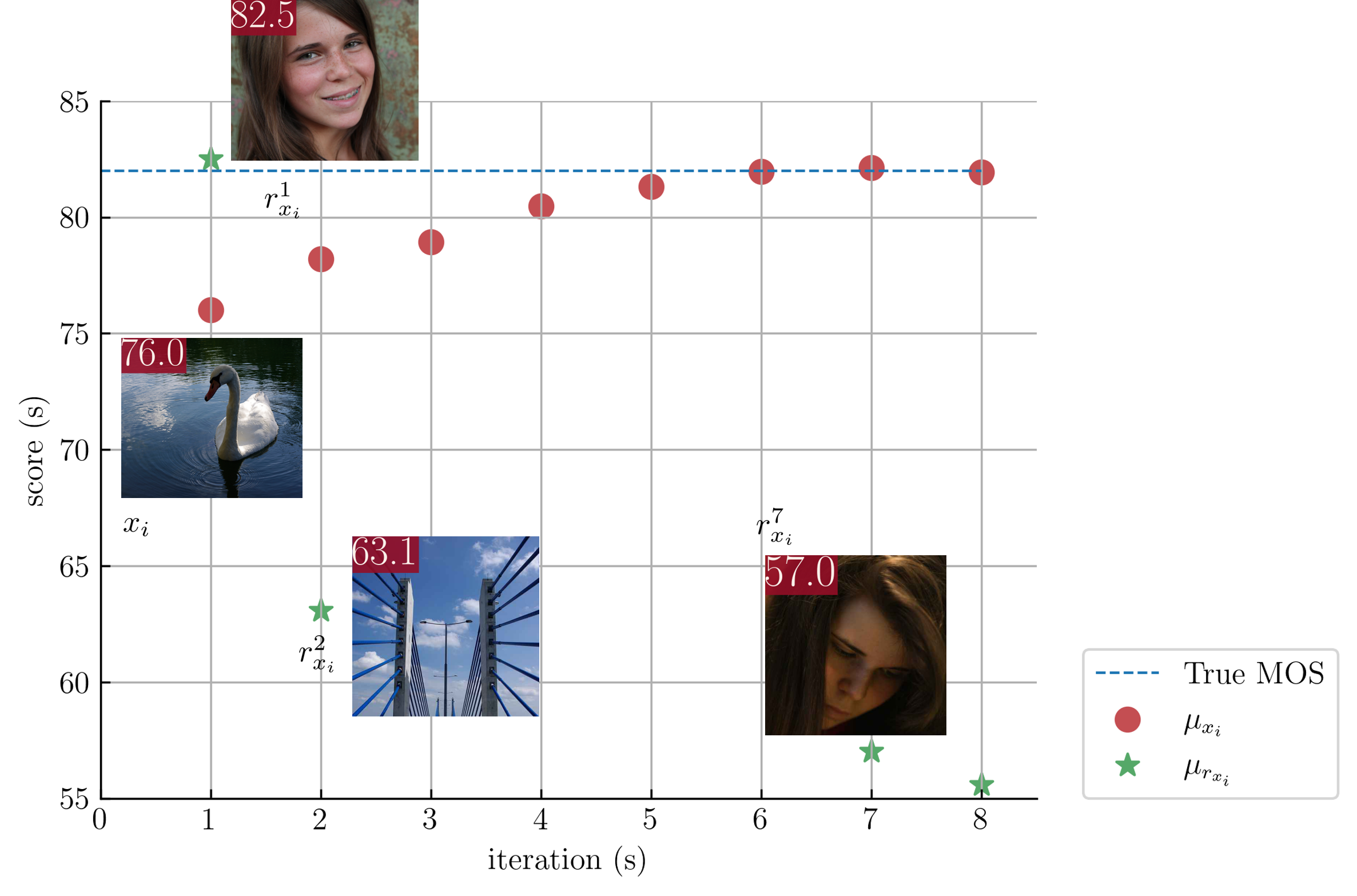} 	
 \caption{The variation of estimated MOS in the iterative update process of PC3.}
	\label{fig:update}
\end{figure}

\begin{table*} 
\centering
\caption{SOS calibration performances on TID, VCL, LIVEC, and KONIQ. We report the median SROCC, PLCC, and MSE results between the calibrated SOS and the ground truth MOS across 10 times of random sampling.}
\label{tab:compare-table}

\scalebox{1.0}{\resizebox{\linewidth}{!}{
\begin{tabular}{c|rrr|rrr|rrr|rrr}
\toprule 
\midrule  
Datasets & \multicolumn{3}{c|}{TID} & \multicolumn{3}{c|}{VCL} & \multicolumn{3}{c|}{LIVEC} & \multicolumn{3}{c}{KONIQ} \\ \midrule
Methods & \multicolumn{1}{c}{SROCC} & \multicolumn{1}{c}{PLCC} & \multicolumn{1}{c|}{MSE} & \multicolumn{1}{c}{SROCC} & \multicolumn{1}{c}{PLCC} & \multicolumn{1}{c|}{MSE} & \multicolumn{1}{c}{SROCC} & \multicolumn{1}{c}{PLCC} & \multicolumn{1}{c|}{MSE} & \multicolumn{1}{c}{SROCC} & \multicolumn{1}{c}{PLCC} & \multicolumn{1}{c}{MSE} \\  
\midrule
MOS & 1.0000 & 1.0000 & 0.0000 & 1.0000 & 1.0000 & 0.0000 & 1.0000 & 1.0000 & 0.0000 & 1.0000 & 1.0000 & 0.0000 \\
SOS & 0.6993 & 0.7048 & 0.0185 & 0.8317 & 0.8422 & 0.0233 & 0.6984 & 0.7228 & 0.0350 & 0.6519 & 0.6916 & 0.0230 \\
PC3 &  0.7683 & 0.7807 & 0.0103 & 0.8834 & 0.8834 & 0.0156 & 0.7509 & 0.7715 & 0.0211 & 0.7237 & 0.7584 & 0.0148 \\
$\Delta$(\%) & \textbf{9.8670} & \textbf{10.7690} & \textbf{-44.3243} & \textbf{6.2162} & \textbf{4.8919} & \textbf{-33.0472} & \textbf{7.5172} & \textbf{6.7377} & \textbf{-39.7143} & \textbf{11.0140} & \textbf{9.6588} & \textbf{-35.6522} \\ 
\midrule
\bottomrule
\end{tabular}}}
\end{table*}

\begin{table*} 
\centering
\caption{SOS calibration performance on TID, VCL, LIVEC, and KONIQ under different bias rates. We report the median SROCC, PLCC, and MSE results between the calibrated SOS and the ground truth MOS across 10 times of random sampling.}
\label{tab:rate-table}
\scalebox{1.0}{\resizebox{\linewidth}{!}{
\begin{tabular}{cc|rrr|rrr|rrr|rrr}
\toprule 
\midrule 
Datasets &  & \multicolumn{3}{c|}{TID} & \multicolumn{3}{c|}{VCL} & \multicolumn{3}{c|}{LIVEC} & \multicolumn{3}{c}{KONIQ} \\ \midrule
Methods & rate & \multicolumn{1}{c}{SROCC} & \multicolumn{1}{c}{PLCC} & \multicolumn{1}{c|}{MSE} & \multicolumn{1}{c}{SROCC} & \multicolumn{1}{c}{PLCC} & \multicolumn{1}{c|}{MSE} & \multicolumn{1}{c}{SROCC} & \multicolumn{1}{c}{PLCC} & \multicolumn{1}{c|}{MSE} & \multicolumn{1}{c}{SROCC} & \multicolumn{1}{c}{PLCC} & \multicolumn{1}{c}{MSE} \\  
\midrule
SOS & \multirow{2}{*}{0.6} & 0.7959 & 0.7923 & 0.0109 & 0.8924 & 0.8966 & 0.0141 & 0.7799 & 0.7985 & 0.0225 & 0.7490 & 0.7836 & 0.0139 \\
PC3 &  & 0.8343 & 0.8474 & 0.0064 & 0.9151 & 0.9121 & 0.0113 & 0.7998 & 0.8226 & 0.0154 & 0.8012 & 0.8277 & 0.0093 \\
$\Delta$(\%) &  & \textbf{4.8247} & \textbf{6.9544} & \textbf{-41.2844} & \textbf{2.5437} & \textbf{1.7288} & \textbf{-19.8582} & \textbf{2.5516} & \textbf{3.0182} & \textbf{-31.5556} & \textbf{6.9693} & \textbf{5.6279} & \textbf{-33.0935}  \\ \midrule
SOS & \multirow{2}{*}{0.8} & 0.7522 & 0.7533 & 0.0144 & 0.8586 & 0.8604 & 0.0191 & 0.7340 & 0.7528 & 0.0299 & 0.6977 & 0.7352 & 0.0187 \\
PC3 &  & 0.8158 & 0.8321 & 0.0075 & 0.8912 & 0.8868 & 0.0142 & 0.7722 & 0.7900 & 0.0189 & 0.7676 & 0.7951 & 0.0119 \\ 
$\Delta$(\%) &  & \textbf{8.4552} & \textbf{10.4606} & \textbf{-47.9167} & \textbf{3.7969} & \textbf{3.0683} & \textbf{-25.6545} & \textbf{5.2044} & \textbf{4.9416} & \textbf{-36.7893} & \textbf{10.0186} & \textbf{8.1474} & \textbf{-36.3636} \\ 
\midrule
\bottomrule
\end{tabular}}}
\end{table*}

\begin{figure}
    \centering
    \includegraphics[width=1\linewidth]{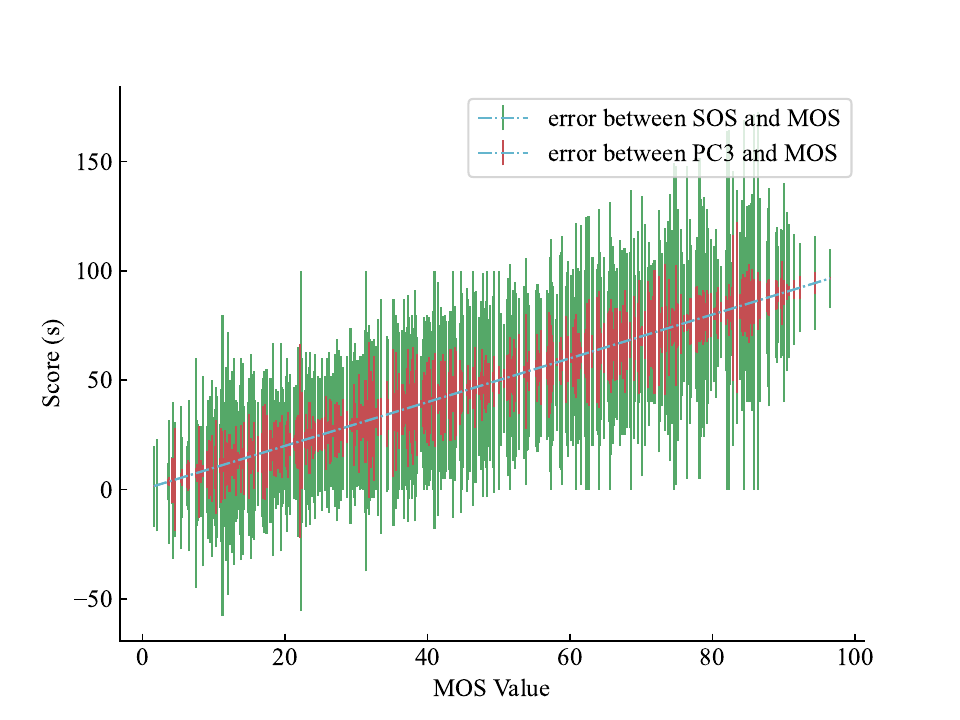}    
    \caption{Error bar between MOS and SOS/PC3 in VCL dataset.}
    \label{fig:case1}
    \vspace{-0.9em}
\end{figure} 

\begin{table*} 
\centering
\caption{The performance of IQA model learned from different subjective quality labels. We report the median
SROCC, PLCC, and MSE results between the predicted MOS of NIMA and the ground truth MOS across 10 times of random splitting.}
\label{tab:iqa-table}

\scalebox{1.0}{\resizebox{\linewidth}{!}{
\begin{tabular}{c|rrr|rrr|rrr|rrr}
\toprule 
\midrule 
Datasets & \multicolumn{3}{c|}{TID} & \multicolumn{3}{c|}{VCL} & \multicolumn{3}{c|}{LIVEC} & \multicolumn{3}{c}{KONIQ} \\ \midrule
Methods & \multicolumn{1}{c}{SROCC} & \multicolumn{1}{c}{PLCC} & \multicolumn{1}{c|}{MSE} & \multicolumn{1}{c}{SROCC} & \multicolumn{1}{c}{PLCC} & \multicolumn{1}{c|}{MSE} & \multicolumn{1}{c}{SROCC} & \multicolumn{1}{c}{PLCC} & \multicolumn{1}{c|}{MSE} & \multicolumn{1}{c}{SROCC} & \multicolumn{1}{c}{PLCC} & \multicolumn{1}{c}{MSE} \\  
 \midrule
MOS & 0.7297 & 0.8011 & 0.0261 & 0.9293 & 0.9233 & 0.0130 & 0.8110 & 0.8538 & 0.0120 & 0.8937 & 0.9126 & 0.0032 \\
SOS & 0.4677 & 0.4956 & 0.0443 & 0.7620 & 0.7600 & 0.0357 & 0.5515 & 0.5630 & 0.0528 & 0.5577 & 0.5838 & 0.0271 \\
PC3 & 0.5134 & 0.5605 & 0.0318 & 0.8589 & 0.8633 & 0.0203 & 0.6787 & 0.6883 & 0.0283 & 0.6835 & 0.7109 & 0.0140 
\\
$\Delta$(\%) & \textbf{9.7712} & \textbf{13.0952} & \textbf{-28.2167} & \textbf{12.7165} & \textbf{13.5921} & \textbf{-43.1373} & \textbf{23.0644} & \textbf{22.2558} & \textbf{-46.4015} & \textbf{22.5569} & \textbf{21.7712} & \textbf{-48.3395} \\  
\midrule
\bottomrule
\end{tabular}}}
\end{table*}

\subsection{Performance of SOS Calibration} 
In Fig. \ref{fig:update}, we first illustrate the variation of estimated MOS in the iterative update process, where an input image $x_i$ and its reference images in three different iterations are visualized for clarity. It is seen that our calibrated result $\mu_{x_n}$ gradually approaches its ground-truth MOS, and the change of the reference images would not affect the stability of the iterative update process. 

The overall SOS calibration performances across all databases are shown in Table \ref{tab:compare-table}. We can find that the proposed PC3 is effective in calibrating the biased SOS, which significantly improves the SRCC, PLCC, and MSE performance across all databases. In addition, the proposed method is also robust to different distortion types, whose performance improvements are similar to the synthetic distortion in TID, VCL, and the authentic distortions in LIVEC, KONIQ. For clarity, in Fig. \ref{fig:case1}, we also show the error bar between MOS and SOS/PC3 across the full range of subjective qualities in the VCL database. It is seen that the proposed PC3 could reduce the bias of SOS across different subjective qualities.

To further investigate the impact of PC3 on subjective labels mixed by MOS and SOS, we introduce the bias rate to control the proportion of SOS on the whole subjective label set. As shown in Table \ref{tab:rate-table}, the proposed method is also efficient for calibrating the mixed subjective labels, whose performances are superior to the SOS. Meanwhile, we can find that the performance improvement of PC3 is more significant for a high bias rate. It is not surprising that a higher bias rate provides a larger improvement space from more SOS.

\subsection{Impact for IQA Model Learning} 
To investigate the impact of PC3 on IQA model learning, we train a representative deep IQA model NIMA~\cite{talebi2018nima} with the ground-truth MOS, SOS, and the calibrated SOS via PC3, respectively. It is seen that the performance of NIMA would significantly degrade when SOS is directly used as the subjective label. By contrast, with the calibration of our PC3, the performance degradation of NIMA would be efficiently alleviated.  

\subsection{Generalization to Few Opinion Scores}
\begin{figure}
    \centering
    \includegraphics[width=0.8\linewidth]{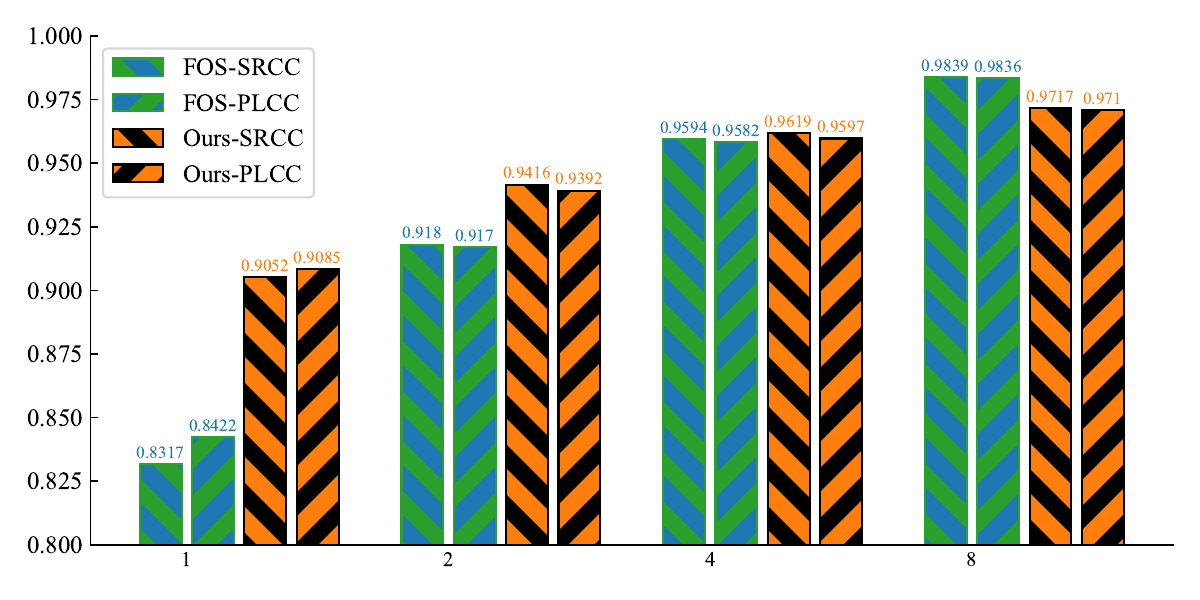}    
    \caption{Performance comparison of FOS with different numbers of subjects.}
    \label{fig:case2}
    \vspace{-0.9em}
\end{figure}

Besides SOS, we further investigate the generalization ability of the proposed PC3 on the mean value of Few Opinion Scores (FOS), which contain multiple subjects but the number of subjects is smaller than the minimal requirement, i.e., 15, in \cite{installationsitu,union2016methods,bt2020methodologies,li2020simple}. We conduct this experiment on VCL which provides all raw opinion scores for each image. The investigated number of subjects includes 1, 2, 4, and 8, and the proposed PC3 is applied to the mean values of FOS to calibrate the bias. Fig. \ref{fig:case2} shows the comparison results between the mean value of FOS and our PC3. It is seen that the PC3 outperforms FOS when the number of subjects is smaller than 4. Although PC3 may degrade the FOS when the number of subjects increases to 8, our performance is comparable, and the reported SRCC and PLCC results are both very high, which are larger than 0.97.   

\begin{table}
\centering
\caption{Performance comparison of PC3 with different $\alpha$ settings. We conduct this test on the VCL database and report the results between the calibrated SOS and the ground-truth MOS.}
\label{tab:weight-table} 
\scalebox{1.1}{\begin{tabular}{l|c|c|c|c}
\toprule 
\midrule 
 & SROCC & \multicolumn{1}{l|}{PLCC} & \multicolumn{1}{l|}{KROCC} & \multicolumn{1}{l}{MSE} \\ \midrule
$\alpha=0.8$ & 0.6792 & 0.6822 & 0.4946 & 0.0416 \\ \midrule
$\alpha=0.6$ & 0.7814 & 0.7765 & 0.5923 & 0.0298 \\ \midrule
$\alpha=0.2$ & 0.8934 & 0.8911 & 0.7104 & 0.0150 \\ \midrule
$\alpha=0.1$ & 0.9052 & 0.9085 & 0.7286 & 0.0124 \\ \midrule
$\alpha=0.0$ & 0.8317 & 0.8422 & 0.6436 & 0.0233 \\ \midrule
\bottomrule
\end{tabular}}
\end{table} 

\subsection{Effect of $\alpha$}
This section investigates the robustness of PC3 toward different $\alpha$ settings on the VCL database. As shown in Eq. (10), $\alpha$ controls the update speed of SOS. Too small or too large $\alpha$ may result in under- or over-calibration. As shown in Table \ref{tab:weight-table}, the PC3 is efficient when $\alpha$ is smaller than 0.2 and significantly degrades when $\alpha$ is larger than 0.6. The best calibration results were obtained in $\alpha=0.1$, which is employed by the proposed method.  

\section{Conclusions}
 
This paper proposes a simple yet efficient subjective quality label post-processing method for IQA. Instead of relying on the bias and inconsistency analysis across multiple subjects, we only utilize the SOS and the perceptual correlation of multiple images, which significantly reduces the cost of MOS collection. Following the normal distribution assumption toward SOS and the perceptual constancy constraint, we formulate the MOS estimation as a constrained maximum likelihood estimation problem, which could be solved via an alternative optimization strategy. Extensive experiments verify the effectiveness of the proposed method for SOS calibration and IQA model learning. Meanwhile, the proposed method also shows a powerful generalization ability to the post-processing of a few opinion scores. 

\bibliographystyle{IEEEtran}
\bibliography{wl}
\end{document}